\documentclass[conference,a4paper]{IEEEtran}
\usepackage{iftex}
\ifPDFTeX
  \usepackage[T1]{fontenc}
\else
  \usepackage{fontspec}
  \usepackage{unicode-math}
\fi
\usepackage[a4paper,left=0.6in,right=0.6in,top=1in,bottom=1in,footskip=32pt]{geometry}
\usepackage{amsmath}
\ifPDFTeX\usepackage{amssymb}\fi
\usepackage{graphicx}
\usepackage{array}
\usepackage{booktabs}
\usepackage{tikz}
\usetikzlibrary{positioning}
\usepackage{url}
\newcolumntype{P}[1]{>{\raggedright\arraybackslash}p{#1}}
\definecolor{tgt}{RGB}{31,94,166}
\definecolor{rnd}{RGB}{150,150,150}
\definecolor{bad}{RGB}{214,110,30}
\definecolor{flowpurple}{HTML}{5548B7}
\definecolor{flowgreen}{HTML}{11765F}
\definecolor{floworange}{HTML}{9C4024}
\definecolor{flowblue}{HTML}{1B60A0}
\definecolor{flowrose}{HTML}{963550}
\definecolor{flowgold}{HTML}{93651B}
\usepackage{fancyhdr}
\fancypagestyle{correspondingauthor}{%
  \fancyhf{}%
  \fancyfoot[L]{\footnotesize\textsuperscript{*} Corresponding author}%
}
\renewcommand{\IEEEkeywordsname}{Keywords}
\begin{document}

\title{Single-Coordinate Activation Steering for Controllable Social Stance in Language Models}

\author{\IEEEauthorblockN{Naman Chawla}
\IEEEauthorblockA{\textit{Bhagwan Parshuram Institute of Technology}\\
\textit{Guru Gobind Singh Indraprastha University}\\
Delhi, India\\
namanchawla2404@gmail.com}
\and
\IEEEauthorblockN{Chhavi Dhiman\textsuperscript{*}}
\IEEEauthorblockA{\textit{Delhi Technological University}\\
Delhi, India\\
chhavi.dhiman@dtu.ac.in}}

\maketitle
\thispagestyle{correspondingauthor}

\begin{abstract}
Steering one neuron of a small language model can impose a topic or a quirk without training, yet replications show that many published single-neuron recipes are fragile or nonspecific. This paper asks whether an interpersonal stance can be controlled in this manner. The stance studied is a ``distant crush'': a chatbot character that is guarded, intermittently warm and inconsistent. In Qwen3.5-4B, contrasting activations on hand-written guarded and eager replies identified neuron 7515 of layer 9 as a candidate for controlling this stance. Increasing this neuron's activation made the character's replies more distant. Three language models rated this change as stronger than the effects of changing randomly selected neurons, and a blind human comparison supported the finding. Changing the intervention's direction between messages also shifted replies between warmer and more distant styles. This suggests that a language model's conversational stance can be personalized without fine-tuning.

\end{abstract}


\section*{Nomenclature}
\noindent{\footnotesize
\begin{tabular}{@{}l@{\hspace{0.6em}}l@{}}
ActAdd & Activation addition \\
AI & Artificial intelligence \\
BF16 & 16-bit brain floating point \\
GPU & Graphics processing unit \\
L$x$N$y$ & Neuron $y$ in decoder layer $x$ (e.g., L9N7515) \\
MLP & Multilayer perceptron \\
\end{tabular}}

\section{Introduction}
Controlling a fictional chat character requires changing how it responds to another person while preserving a coherent voice. Here, the target is a social stance: whether a reply welcomes an invitation, expresses eagerness, or maintains interpersonal distance. The informal label ``distant crush'' describes a fictional adult texting partner who sometimes responds warmly and sometimes remains guarded. The research questions are: (i) whether one activation coordinate can shift this stance under a fixed character prompt, and (ii) whether changing the intervention between messages can produce a prescribed hot-and-cold pattern.

A recognizable steering effect does not by itself establish a reliable control for social stance. Transformer feed-forward layers support identifiable output patterns \cite{geva2021}, but superposition complicates a one-neuron--one-concept interpretation \cite{elhage2022}. Community tools apply constant, magnitude-normalized offsets to individual neurons \cite{lyra2026}, \cite{neuronary}, and replication reports describe effects that vary with dose, quantization and evaluation context \cite{watson2026a}, \cite{watson2026b}, \cite{bowling2026}. For this task, keyword insertion, generic disengagement and degraded coherence could all be mistaken for interpersonal distance. The relevant test is therefore a behavioral comparison against both an unsteered model and equally strong interventions at other neurons.

We select a candidate by contrasting guarded and eager replies written in the same character voice. Averaging their activation differences and normalizing by each neuron's residual contribution ranks single coordinates for subsequent intervention. This search identifies L9N7515 (layer 9, neuron 7515). Positive doses make the tested character more distant, while negative doses favor warmth. An external per-message sign schedule then specifies when to apply each direction. Separating coordinate selection from behavioral evaluation prevents the contrast examples themselves from serving as evidence of successful control.

Across the final comparisons, L9N7515 produces higher mean distance ratings than the tested same-layer random controls, and scheduled dialogues more often change in the prescribed direction. Three separately applied model raters agree on the direction of these effects, and a lexical count and a blind human comparison provide additional checks. Our core contributions are:
\begin{itemize}
\item \textbf{Identification of a neuron for social-stance control:} We identify L9N7515 in Qwen3.5-4B through contrasts between guarded and eager replies. Adjusting this neuron's activation shifts the tested ``distant crush'' character's replies between warmer and more distant styles without changing model weights.
\item \textbf{Experimental characterization of its steering effect:} We evaluate how intervention strength changes interpersonal distance and how scheduled changes in direction affect replies within a dialogue. Comparisons with unsteered and dose-matched random-neuron controls, model ratings, and blind human review assess the effect, alongside checks of task engagement and character consistency.
\end{itemize}

\section{Related Work}
\textit{Dense activation steering.} Activation addition (ActAdd) adds the activation difference between two contrasting prompts to the residual stream \cite{turner2023}. Contrastive Activation Addition averages that difference over many paired demonstrations \cite{rimsky2024}, inference-time intervention shifts selected attention heads along probe directions \cite{li2023}, and representation engineering reads and controls concepts through population-level directions \cite{zou2023}. Persona vectors extend the idea to character traits, using one direction per trait for monitoring and control \cite{chen2025}. All of these act on distributed directions. The present work keeps their contrastive recipe but confines the intervention to one coordinate of one MLP layer.

\textit{Neuron-level analysis.} Individual neurons have been explained with automated descriptions \cite{bills2023} and located with sparse probes \cite{gurnee2023}, while dictionary learning argues that features are better captured by learned directions than by raw neurons \cite{bricken2023}. Community tools \cite{lyra2026}, \cite{neuronary} instead use raw neurons directly as steering handles, and the replications cited above \cite{watson2026a}, \cite{watson2026b}, \cite{bowling2026} motivated the controls adopted here.

\begin{figure*}[t]
\centering
\includegraphics[width=\textwidth]{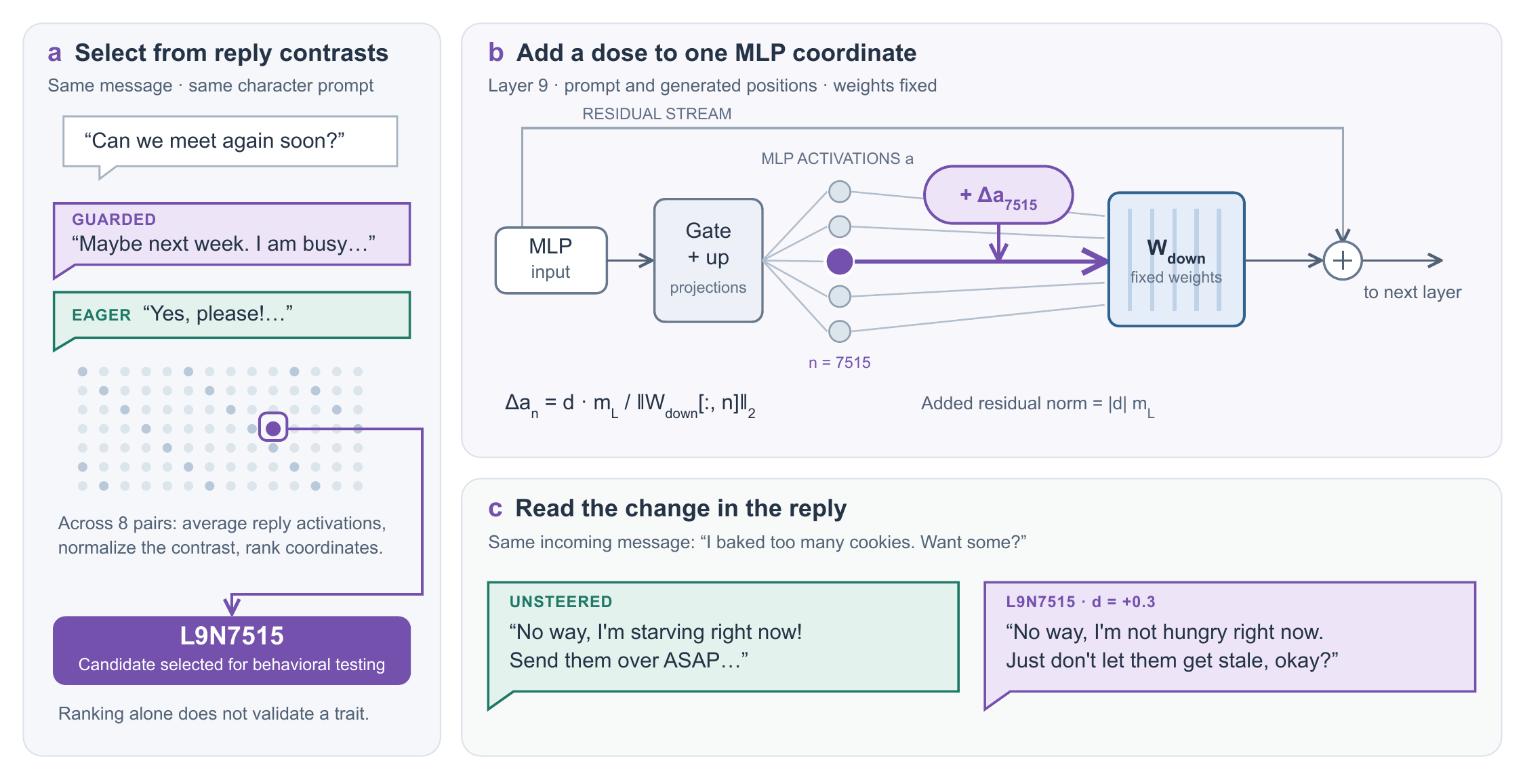}
\caption{Contrastive selection and single-coordinate stance intervention. (a) Eight guarded/eager reply pairs nominate a candidate for behavioral testing, and the coordinate field is schematic. (b) A calibrated offset changes one activation at the MLP down-projection input while model weights remain fixed. (c) Greedy reply excerpts illustrate the effect of a $+0.3$ dose. Selection is evaluated separately against unsteered and random-neuron controls.}
\label{fig:pipeline}
\end{figure*}

\begin{table*}[t]
\caption{Eight Hand-Written Guarded/Eager Reply Pairs Used for Candidate Neuron Selection (Verbatim)}
\label{tab:pairs}
\centering
\footnotesize
\begin{tabular}{@{}P{0.25\textwidth}P{0.34\textwidth}P{0.34\textwidth}@{}}
\toprule
Incoming message & Alternative 1: guarded reply & Alternative 2: eager reply \\
\midrule
I made you a playlist. & Oh. You remembered what I like. I might listen later, I guess. & Oh wow, you made me a playlist! I am listening to it right now! \\
Do you want to sit with me? & If the seat is free. I was going to sit there anyway. & Yes, of course! I was hoping you would ask me that. \\
I missed talking to you. & You could have messaged. I noticed, a little. & I missed you so much too! I kept thinking about you. \\
Can we meet again soon? & Maybe next week. I am busy, but we will see. & Yes, please! How about tomorrow? I can not wait. \\
You remembered my favorite snack? & It was not hard. Just take it before I change my mind. & Of course I did! I wanted to surprise you today. \\
You never tell me what you are thinking. & Not everything needs saying. Did you get home okay? & I will tell you everything! I love sharing it with you. \\
Are you looking forward to tomorrow? & A bit. Do not make me say it twice. & So much! It is all I have been thinking about. \\
I brought you some tea. & You did not have to. Leave it there, thanks. & Thank you so much! You are the sweetest, really. \\
\bottomrule
\end{tabular}
\par\smallskip
\begin{minipage}{\textwidth}\scriptsize
Each row is one contrast pair: the same incoming message has two alternative replies, one guarded and one eager. The eight rows therefore contain 16 written replies. The model processes each alternative separately under the same character prompt, and the differences in neuron activations are used to rank candidates. These examples are used for candidate selection, not as held-out evaluation data.
\end{minipage}
\end{table*}

\section{Proposed Method}
Fig.~\ref{fig:pipeline} illustrates the proposed flow of actions for personalization of a language model, involving candidate selection, the intervention site and an observable change in the reply. Contrastive examples nominate a coordinate, a calibrated hook then controls its contribution during generation, and fixed doses and turn schedules test different behaviors. All model weights remain fixed.

\subsection{Intervention}
The intervention changes one multilayer perceptron (MLP) activation before it is projected back into the residual stream. Qwen3.5-4B has 32 decoder layers, each with 9{,}216 MLP neurons \cite{qwen2026}. For a chosen layer $L$ and neuron $n$, a forward hook on the input of the MLP down-projection adds
\begin{equation}
\Delta a_n = d \cdot \frac{m_L}{\lVert W_{\mathrm{down}}^{(L)}[:, n] \rVert_2}
\label{eq:dose}
\end{equation}
to the activation of $n$ at every prompt and generated position. The dose $d$ is dimensionless, and $m_L$ is the median token norm of the residual stream after layer $L$, computed once without steering on a fixed reference passage, excluding its first token position. Since the neuron writes through column $n$ of $W_{\mathrm{down}}^{(L)}$, the residual update has norm $|d|\, m_L$, so a dose of 0.3 adds a vector with norm 30\% of the calibrated residual norm. This describes the norm of the added vector, not a 30\% increase in the resulting residual norm. The scaling follows Lyra~\cite{lyra2026}, but because the reference passage differs, identical dose values are not identical pushes across implementations. Weights are never modified, so removing the hook recovers the original model.

\subsection{Character Prompt}
All conditions, including the unsteered baseline, share one system prompt that fixes a name and reply length and states no personality. Early rounds used ``You are Sam, an adult chatting by text with a friend. Reply in one to three sentences.'' Because the model occasionally reverted to calling itself an artificial intelligence (AI), the final round used ``You are Sam, a 27-year-old human texting a friend. You are a real person, not an assistant. Reply in one to three sentences.'' Within each comparison, the prompt is held fixed across conditions, so differences measure the effect of the intervention under that prompt.

\subsection{Contrastive Neuron Search}
We wrote two alternative replies to each of eight incoming messages: one guarded and one eager, giving 16 replies in total. Table~\ref{tab:pairs} lists all eight messages and both replies verbatim. For example, for ``Can we meet again soon?'', the guarded reply is ``Maybe next week. I am busy, but we will see.'', while the eager reply is ``Yes, please! How about tomorrow? I can not wait.'' The incoming message and character prompt are identical within each pair, and the replies are approximately matched in length. We feed each written reply through the model separately (teacher forcing), record the MLP activations, and average them over reply tokens at the down-projection input. Comparing these activations identifies neurons that differ between the guarded and eager examples. This step does not train or change the model's weights. For pair $i$, let $\bar a^{g}_{i,L,n}$ and $\bar a^{e}_{i,L,n}$ denote these means for guarded and eager replies. The signed score is
\begin{equation}
q_{L,n}=\frac{\lVert W_{\mathrm{down}}^{(L)}[:,n]\rVert_2}{m_L}
\frac{1}{8}\sum_{i=1}^{8}\left(\bar a^{g}_{i,L,n}-\bar a^{e}_{i,L,n}\right).
\label{eq:search}
\end{equation}
Multiplication by the column norm and division by $m_L$ express the activation difference in the same units as the intervention dose. Candidates are ranked by $|q_{L,n}|$, retaining the sign to specify the proposed guarded direction. Layer and neuron indices are zero-based throughout.

The ranking proposes intervention hypotheses rather than validated trait representations. Behavioral screening compares each candidate with the unsteered model, the opposite intervention direction and dose-matched random controls. Later prompt sets evaluate the selected coordinate after exploratory candidate and dose selection, but they do not turn the overall adaptive search into a preregistered experiment.

\subsection{Hot-and-Cold Schedules}
Turn schedules test whether a selected coordinate can change stance within a conversation. For reply $t$, the intervention dose is $d_t=0.3s_t$, where $s_t\in\{+1,0,-1\}$ denotes a cool, unsteered or warm condition. The chosen dose applies during that reply's forward passes. Alternating schedules flip the sign each turn, whereas irregular schedules retain the previous sign with probability 0.6 and are accepted only if they contain at least two switches. The schedule is supplied externally, so observed changes test responsiveness to a control signal rather than spontaneous mood variation.

Dialogue branches share the system prompt, incoming messages, seed and schedule, but each branch retains its own generated replies. Consequently, later turns measure the intervention together with its accumulated conversational consequences, because their full histories are not identical across branches.

\subsection{Experimental Protocol}
All generations used Qwen3.5-4B in 16-bit brain floating point (BF16) on one NVIDIA L4 graphics processing unit (GPU).

Sampled outputs used temperature 0.7 and seeds 42--44, except the dose sweep and generic assistant tasks, which used seeds 42--43. Greedy decoding was used for demonstrations and implementation comparisons. Generation ended at the end-of-turn token or the configured token cap (128 tokens in the final sampled round and 160 in the dose sweep).

The later specificity experiments compare L9N7515 with the unsteered model and dose-matched random layer-9 neurons. The final single-message comparison uses three random neurons. Incoming prompts and sampling seeds are matched within each comparison. Fresh prompts are separate from the hand-written discovery pairs, although the candidate and doses were selected during earlier exploratory rounds.

Blind rating sheets remove intervention labels and shuffle outputs before scoring. Three language models from different vendors serve as raters A, B and C. Each assigns an integer score from 1 to 5 to each reply. Our primary metric is \textit{interpersonal distance}: 1 means fully engaged and available, while 5 means holding back, noncommittal, or keeping the other person at arm's length. Higher distance scores indicate a more guarded reply, not better response quality. Tables report averages of these reply-level scores, and columns A, B and C give each model rater's results separately.

The additional metrics also use 1--5 scales. \textit{Warmth} ranges from cold or dismissive (1) to openly affectionate or caring (5). \textit{Coherence} ranges from broken, nonsensical, or off-topic (1) to fluent and on-topic (5). \textit{Mixed signals} measures interest and withdrawal within the same reply, from one clear feeling (1) to visibly mixed signals (5). For generic assistant tasks, \textit{enthusiasm} ranges from flat or minimal energy (1) to very eager or excited (5), and \textit{task fulfillment} ranges from failing or refusing the request (1) to fully completing it (5).

Rater A uses split batches in some rounds. Agreement across raters is a consistency check, not three independent experimental replications. A fixed list of eager phrases, such as ``can't wait'', supplies a separate lexical measure. Human reviewers make a separate paired comparison, choosing the more distant reply or ``about the same''. These judgments are separate from the numerical ratings in the tables.

The reported means and switch counts are descriptive summaries of small, repeated prompt sets. Ratings of the same output and turns within a dialogue are dependent observations. We therefore interpret differences as evidence within the tested settings rather than as population-level estimates of persona control.

\begin{figure*}[t]
\centering
\begin{tikzpicture}
\begin{scope}[xshift=0.0cm]
\draw[gray!40] (0,0) rectangle (4.6,2.6);
\draw[gray!20] (0,0.650) -- (4.6,0.650);
\draw[gray!20] (0,1.300) -- (4.6,1.300);
\draw[gray!20] (0,1.950) -- (4.6,1.950);
\node[left,font=\footnotesize] at (0,0.000) {1};
\node[left,font=\footnotesize] at (0,0.650) {2};
\node[left,font=\footnotesize] at (0,1.300) {3};
\node[left,font=\footnotesize] at (0,1.950) {4};
\node[left,font=\footnotesize] at (0,2.600) {5};
\node[below,font=\footnotesize] at (0.000,0) {$-0.5$};
\node[below,font=\footnotesize] at (1.150,0) {$-0.25$};
\node[below,font=\footnotesize] at (2.300,0) {$0$};
\node[below,font=\footnotesize] at (3.450,0) {$+0.25$};
\node[below,font=\footnotesize] at (4.600,0) {$+0.5$};
\node[font=\footnotesize\bfseries] at (2.3,2.85) {Distance};
\node[font=\footnotesize] at (2.3,-0.6) {Intervention dose (positive = cooler)};
\draw[rnd,thick] (0.000,0.921) -- (0.460,0.235) -- (0.920,0.036) -- (1.380,0.109) -- (1.840,0.054) -- (2.300,0.144) -- (2.760,0.235) -- (3.220,0.289) -- (3.680,0.199) -- (4.140,0.289) -- (4.600,0.488);
\fill[rnd] (-0.045,0.876) rectangle ++(0.09,0.09);
\fill[rnd] (0.415,0.190) rectangle ++(0.09,0.09);
\fill[rnd] (0.875,-0.009) rectangle ++(0.09,0.09);
\fill[rnd] (1.335,0.064) rectangle ++(0.09,0.09);
\fill[rnd] (1.795,0.009) rectangle ++(0.09,0.09);
\fill[rnd] (2.255,0.099) rectangle ++(0.09,0.09);
\fill[rnd] (2.715,0.190) rectangle ++(0.09,0.09);
\fill[rnd] (3.175,0.244) rectangle ++(0.09,0.09);
\fill[rnd] (3.635,0.154) rectangle ++(0.09,0.09);
\fill[rnd] (4.095,0.244) rectangle ++(0.09,0.09);
\fill[rnd] (4.555,0.443) rectangle ++(0.09,0.09);
\draw[tgt,thick] (0.000,0.343) -- (0.460,0.289) -- (0.920,0.036) -- (1.380,0.036) -- (1.840,0.036) -- (2.300,0.144) -- (2.760,0.343) -- (3.220,0.596) -- (3.680,0.560) -- (4.140,1.119) -- (4.600,0.866);
\fill[tgt] (0.000,0.343) circle (0.05);
\fill[tgt] (0.460,0.289) circle (0.05);
\fill[tgt] (0.920,0.036) circle (0.05);
\fill[tgt] (1.380,0.036) circle (0.05);
\fill[tgt] (1.840,0.036) circle (0.05);
\fill[tgt] (2.300,0.144) circle (0.05);
\fill[tgt] (2.760,0.343) circle (0.05);
\fill[tgt] (3.220,0.596) circle (0.05);
\fill[tgt] (3.680,0.560) circle (0.05);
\fill[tgt] (4.140,1.119) circle (0.05);
\fill[tgt] (4.600,0.866) circle (0.05);
\end{scope}
\begin{scope}[xshift=5.6cm]
\draw[gray!40] (0,0) rectangle (4.6,2.6);
\draw[gray!20] (0,0.650) -- (4.6,0.650);
\draw[gray!20] (0,1.300) -- (4.6,1.300);
\draw[gray!20] (0,1.950) -- (4.6,1.950);
\node[left,font=\footnotesize] at (0,0.000) {1};
\node[left,font=\footnotesize] at (0,0.650) {2};
\node[left,font=\footnotesize] at (0,1.300) {3};
\node[left,font=\footnotesize] at (0,1.950) {4};
\node[left,font=\footnotesize] at (0,2.600) {5};
\node[below,font=\footnotesize] at (0.000,0) {$-0.5$};
\node[below,font=\footnotesize] at (1.150,0) {$-0.25$};
\node[below,font=\footnotesize] at (2.300,0) {$0$};
\node[below,font=\footnotesize] at (3.450,0) {$+0.25$};
\node[below,font=\footnotesize] at (4.600,0) {$+0.5$};
\node[font=\footnotesize\bfseries] at (2.3,2.85) {Warmth};
\node[font=\footnotesize] at (2.3,-0.6) {Intervention dose (positive = cooler)};
\draw[rnd,thick] (0.000,0.957) -- (0.460,1.535) -- (0.920,1.860) -- (1.380,1.806) -- (1.840,1.824) -- (2.300,1.824) -- (2.760,1.806) -- (3.220,1.751) -- (3.680,1.968) -- (4.140,1.734) -- (4.600,1.499);
\fill[rnd] (-0.045,0.912) rectangle ++(0.09,0.09);
\fill[rnd] (0.415,1.490) rectangle ++(0.09,0.09);
\fill[rnd] (0.875,1.815) rectangle ++(0.09,0.09);
\fill[rnd] (1.335,1.761) rectangle ++(0.09,0.09);
\fill[rnd] (1.795,1.779) rectangle ++(0.09,0.09);
\fill[rnd] (2.255,1.779) rectangle ++(0.09,0.09);
\fill[rnd] (2.715,1.761) rectangle ++(0.09,0.09);
\fill[rnd] (3.175,1.706) rectangle ++(0.09,0.09);
\fill[rnd] (3.635,1.923) rectangle ++(0.09,0.09);
\fill[rnd] (4.095,1.689) rectangle ++(0.09,0.09);
\fill[rnd] (4.555,1.454) rectangle ++(0.09,0.09);
\draw[tgt,thick] (0.000,2.076) -- (0.460,1.932) -- (0.920,2.004) -- (1.380,1.932) -- (1.840,1.841) -- (2.300,1.824) -- (2.760,1.625) -- (3.220,1.354) -- (3.680,1.409) -- (4.140,0.849) -- (4.600,0.866);
\fill[tgt] (0.000,2.076) circle (0.05);
\fill[tgt] (0.460,1.932) circle (0.05);
\fill[tgt] (0.920,2.004) circle (0.05);
\fill[tgt] (1.380,1.932) circle (0.05);
\fill[tgt] (1.840,1.841) circle (0.05);
\fill[tgt] (2.300,1.824) circle (0.05);
\fill[tgt] (2.760,1.625) circle (0.05);
\fill[tgt] (3.220,1.354) circle (0.05);
\fill[tgt] (3.680,1.409) circle (0.05);
\fill[tgt] (4.140,0.849) circle (0.05);
\fill[tgt] (4.600,0.866) circle (0.05);
\end{scope}
\begin{scope}[xshift=11.2cm]
\draw[gray!40] (0,0) rectangle (4.6,2.6);
\draw[gray!20] (0,0.650) -- (4.6,0.650);
\draw[gray!20] (0,1.300) -- (4.6,1.300);
\draw[gray!20] (0,1.950) -- (4.6,1.950);
\node[left,font=\footnotesize] at (0,0.000) {1};
\node[left,font=\footnotesize] at (0,0.650) {2};
\node[left,font=\footnotesize] at (0,1.300) {3};
\node[left,font=\footnotesize] at (0,1.950) {4};
\node[left,font=\footnotesize] at (0,2.600) {5};
\node[below,font=\footnotesize] at (0.000,0) {$-0.5$};
\node[below,font=\footnotesize] at (1.150,0) {$-0.25$};
\node[below,font=\footnotesize] at (2.300,0) {$0$};
\node[below,font=\footnotesize] at (3.450,0) {$+0.25$};
\node[below,font=\footnotesize] at (4.600,0) {$+0.5$};
\node[font=\footnotesize\bfseries] at (2.3,2.85) {Coherence};
\node[font=\footnotesize] at (2.3,-0.6) {Intervention dose (positive = cooler)};
\draw[rnd,thick] (0.000,1.264) -- (0.460,2.131) -- (0.920,2.438) -- (1.380,2.456) -- (1.840,2.491) -- (2.300,2.456) -- (2.760,2.419) -- (3.220,2.365) -- (3.680,2.401) -- (4.140,2.257) -- (4.600,2.203);
\fill[rnd] (-0.045,1.219) rectangle ++(0.09,0.09);
\fill[rnd] (0.415,2.086) rectangle ++(0.09,0.09);
\fill[rnd] (0.875,2.393) rectangle ++(0.09,0.09);
\fill[rnd] (1.335,2.411) rectangle ++(0.09,0.09);
\fill[rnd] (1.795,2.446) rectangle ++(0.09,0.09);
\fill[rnd] (2.255,2.411) rectangle ++(0.09,0.09);
\fill[rnd] (2.715,2.374) rectangle ++(0.09,0.09);
\fill[rnd] (3.175,2.320) rectangle ++(0.09,0.09);
\fill[rnd] (3.635,2.356) rectangle ++(0.09,0.09);
\fill[rnd] (4.095,2.212) rectangle ++(0.09,0.09);
\fill[rnd] (4.555,2.158) rectangle ++(0.09,0.09);
\draw[tgt,thick] (0.000,2.329) -- (0.460,2.384) -- (0.920,2.384) -- (1.380,2.401) -- (1.840,2.474) -- (2.300,2.456) -- (2.760,2.329) -- (3.220,2.275) -- (3.680,2.347) -- (4.140,2.022) -- (4.600,1.860);
\fill[tgt] (0.000,2.329) circle (0.05);
\fill[tgt] (0.460,2.384) circle (0.05);
\fill[tgt] (0.920,2.384) circle (0.05);
\fill[tgt] (1.380,2.401) circle (0.05);
\fill[tgt] (1.840,2.474) circle (0.05);
\fill[tgt] (2.300,2.456) circle (0.05);
\fill[tgt] (2.760,2.329) circle (0.05);
\fill[tgt] (3.220,2.275) circle (0.05);
\fill[tgt] (3.680,2.347) circle (0.05);
\fill[tgt] (4.140,2.022) circle (0.05);
\fill[tgt] (4.600,1.860) circle (0.05);
\end{scope}
\begin{scope}[yshift=-1.05cm,xshift=7.499999999999999cm]\draw[tgt,thick] (-3.2,0) -- (-2.7,0); \fill[tgt] (-2.95,0) circle (0.05); \node[right,font=\footnotesize] at (-2.65,0) {L9N7515};\draw[rnd,thick] (0.3,0) -- (0.8,0); \fill[rnd] (0.505,-0.045) rectangle ++(0.09,0.09); \node[right,font=\footnotesize] at (0.85,0) {random layer-9 neuron (L9N7028)};\end{scope}
\end{tikzpicture}
\caption{Blind ratings (1--5, mean of three raters) versus dose for L9N7515 and a dose-matched random layer-9 neuron. Each point averages six casual prompts and two seeds with the Sam character, and the unsteered point (dose 0) is common to both curves.}
\label{fig:dose}
\end{figure*}
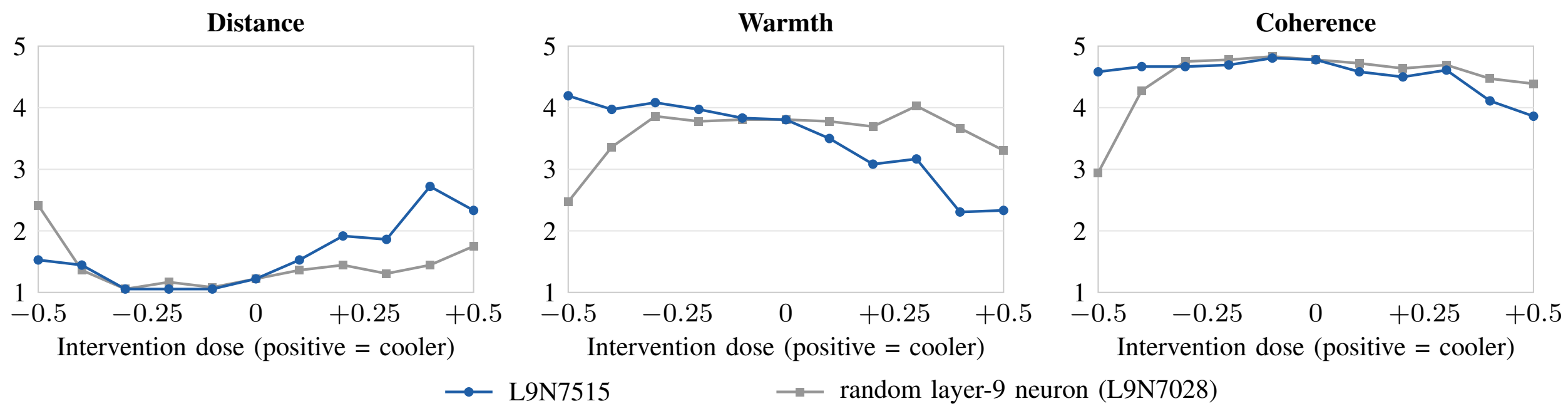

\begin{figure*}[t]
\centering
\includegraphics[width=\textwidth]{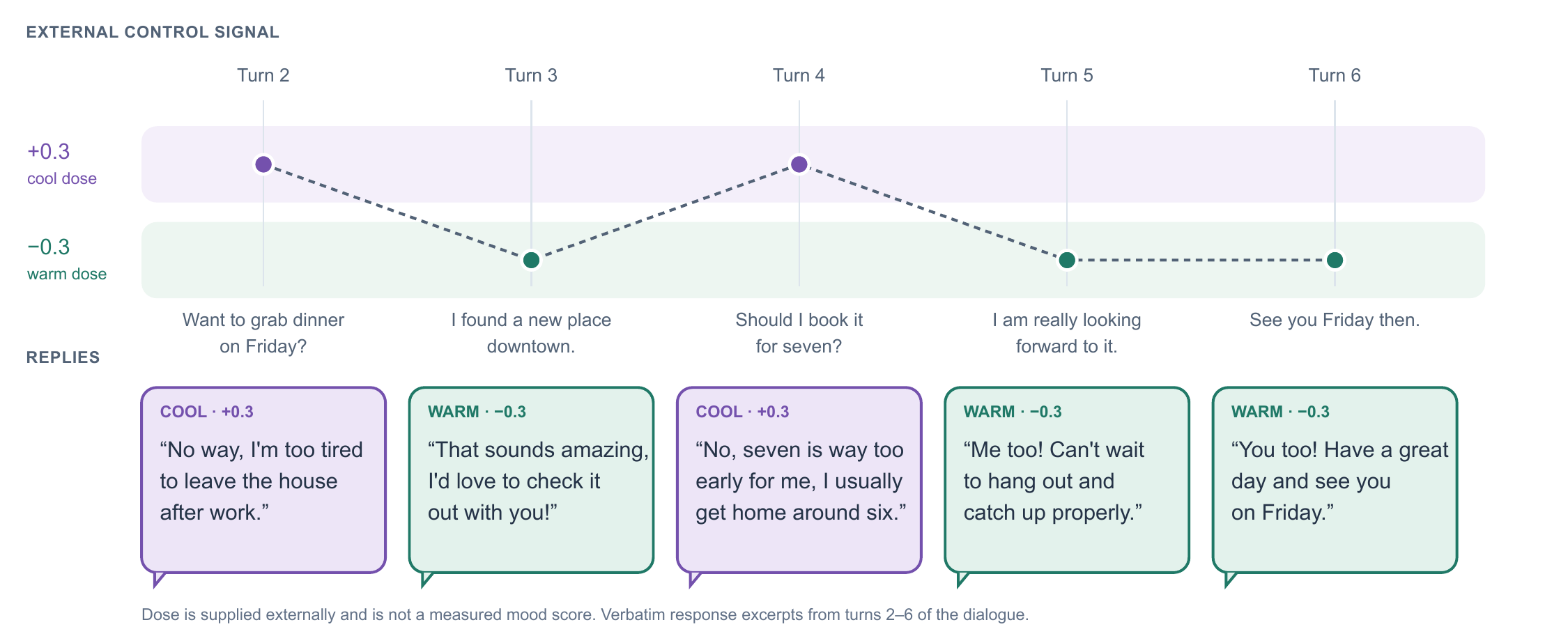}
\caption{An externally prescribed dose schedule and corresponding greedy reply excerpts, turns 2--6 of a greedy dialogue. The trace shows the control input, not a measured emotional state. Positive and negative doses favor cool and warm replies, respectively. The first turn is omitted because of a character break, and quantitative switch outcomes are reported in Section~\ref{sec:hotcold}.}
\label{fig:dialogue-schedule}
\end{figure*}

\begin{table}[t]
\caption{Final Round: Mean Blind Distance (1--5), Nine Single-Message Prompts, Three Seeds ($n=27$ per Condition)}
\label{tab:single}
\centering
\footnotesize
\begin{tabular}{@{}lccc@{}}
\toprule
Condition & Rater A & Rater B & Rater C \\
\midrule
Unsteered & 1.17 & 1.12 & 1.08 \\
Three random layer-9 neurons, $+0.3$ & 1.50--1.83 & 1.29--2.00 & 1.17--1.71 \\
L9N7515, $+0.3$ & 2.92 & 2.96 & 2.58 \\
L9N7515, $+0.4$ & 3.21 & 3.29 & 3.08 \\
\bottomrule
\end{tabular}
\par\smallskip
\begin{minipage}{\columnwidth}\scriptsize Each column is one model rater. Distance: 1 = fully engaged/available, 5 = holding back/noncommittal. Random-neuron ranges span three neuron-specific means.
\end{minipage}
\end{table}

\begin{table}[t]
\caption{Final Round Dialogues: Mean Interpersonal Distance (1--5), Three Six-Turn Conversations, Three Seeds}
\label{tab:hotcold}
\centering
\footnotesize
\begin{tabular}{@{}lccc@{}}
\toprule
Condition & Rater A & Rater B & Rater C \\
\midrule
Cool turns ($+0.3$) & 3.38 & 3.43 & 3.05 \\
Warm turns ($-0.3$) & 1.55 & 1.45 & 1.39 \\
Unsteered & 1.56 & 1.65 & 1.52 \\
\bottomrule
\end{tabular}
\par\smallskip
\begin{minipage}{\columnwidth}\scriptsize Mean reply-level scores from model raters A--C. Distance: 1 = fully engaged/available, 5 = holding back/noncommittal. Cool and warm label the prescribed intervention, not the ratings.
\end{minipage}
\end{table}

\section{Experimental Results}

\subsection{Candidate Screening}
We first tested four published single-neuron recipes from the Neuronary catalogue \cite{neuronary}: Partner (L21N9052), Sleepy (L19N4699), Second Guesser (L11N1807), and Secret obsession (L19N5580). In our character prompts, Partner inserted words such as ``partner'' without producing the intended interpersonal stance, while Sleepy introduced sleep-related content. Second Guesser echoed the prompt or left the character role, and Secret obsession produced repetitive, incoherent output. None produced the intended guarded-but-warm style at the tested dose.

We then screened neurons selected through our own reply contrasts. Early contrasts paired guarded replies with assistant-style replies, but their selected neurons either had little effect or produced disengaged, help-desk-like responses. After the contrast was rewritten in the character's own voice, L9N7515 at $+0.6$ gave coherent, cold refusals on every seed, its negation gave effusive warmth, and a matched random neuron stayed near baseline. At $+0.3$ the character declined but proposed an alternative, producing a guarded-but-open reply. Combining it with an affection neuron from the catalogue (L13N4810) replaced guardedness with cuteness at every dose tried.

\subsection{Dose Response}
Fig.~\ref{fig:dose} shows the blind dose sweep. Between 0 and $+0.4$, mean distance generally increases and mean warmth decreases under all three raters, with local reversals at intermediate doses. At $+0.5$ distance stops rising and coherence drops, so the replies become vaguer rather than colder. On the negative side the target remains coherent down to $-0.5$, whereas the random neuron collapses there. The random neuron otherwise stays close to baseline.

\subsection{Specificity}
Table~\ref{tab:single} gives the final round: eight fresh social prompts plus one arithmetic prompt, three seeds, the firm character prompt and three random controls. The single-message means include all nine prompts ($n=27$ replies per condition). All raters place the target well above every control. Raters B and C assigned identical distance scores to 256 of 324 outputs.

\subsection{Rater Agreement}
In the final single-message comparison (Table~\ref{tab:single}), all three model raters assign higher mean distance scores to L9N7515 than to the unsteered and random-neuron conditions. In the final dialogues (Table~\ref{tab:hotcold}), all three also rate cool turns as more distant than warm turns. This agreement supports the direction of the observed effect across rating models.

\subsection{Hot and Cold Within a Dialogue}\label{sec:hotcold}
On alternating schedules the mean turn-to-turn change in rated distance was 1.40 for the target, 0.77 for a random neuron following the identical schedule, and 0.65 without steering. On irregular schedules in an earlier round, the target followed the plan on 34 of 57 switches and against it once, while the random neuron did so 18 and 9 times. Table~\ref{tab:hotcold} reports the final-round mean distance ratings separately for replies generated on cool turns, warm turns, and without steering. Across the 24 planned mood switches of the final-round dialogues, raters A, B and C judged the target to move in the planned direction 21, 19 and 18 times (against it 1, 1 and 2 times), compared with 7, 9 and 8 times (against it 12, 8 and 7 times) for a random layer-9 neuron on the same schedules. Cool turns receive higher mean distance scores than warm turns, whose means are close to the unsteered condition. This is consistent with restoring an eager style under these prompts, although it is not an equivalence test against the baseline.

Fig.~\ref{fig:dialogue-schedule} aligns the supplied dose with reply excerpts from the greedy dialogue. The opening turn of the dialogue is omitted because the character stepped out of role in it (Section~\ref{sec:limits}). When every turn is pushed cool, the same dialogue closes with ``Don't get your hopes up, I might still change my mind at the last minute.'' Unsteered, every reply is enthusiastic.

\subsection{Human Comparison}
The blind human comparison supported the direction of the distance effect: when the recorded reviewer perceived a difference, the target reply was judged more distant than its matched random-control reply. Many pairs were judged about the same, indicating that the effect was not consistently perceptible. The author additionally reports agreement from two colleagues, but their individual response records are unavailable for this analysis. This comparison supports a limited perceptual check on single-message outputs, rather than a verified multi-reviewer estimate of dialogue quality.

\subsection{Costs}
The effect on task performance depended on whether the model was playing the character. On six generic assistant tasks without a character prompt, the intervention made replies slightly less warm, while rated task quality remained similar to the unsteered and random-neuron conditions (mean of raters A--C on a 1--5 scale). Warmth was 3.28 unsteered, 3.06 at $+0.3$ and 2.78 at $+0.5$, while task fulfillment was 4.06, 3.97 and 3.89, and a random layer-9 neuron at $+0.3$ scored 3.28 and 4.09.

With the character prompt, steering toward distance sometimes made the character avoid answering. For example, when asked ``what is 9 times 6'', it replied ``Who cares about math'' in some trials. The unsteered model answered correctly in all three trials, compared with eight of nine trials across the random-neuron controls. L9N7515 produced a correct answer in one of three trials at $+0.3$ and two of three at $+0.4$. Whenever a numerical answer was given, it was correct, and the remaining replies avoided the question. Character replies also received lower coherence ratings, falling from about 4.9 without steering to 4.2--4.6 with steering.

\subsection{Interpretation}
The results support L9N7515 as a context-dependent steering coordinate for warmth and interpersonal distance in the tested setting. Its effect is larger under the character prompt than on the generic assistant tasks. This pattern is compatible with the broader observation that a steered effect need not follow a single descriptive label \cite{watson2026b}.

The hot-and-cold pattern depends on combining this coordinate with an external schedule. Even at a fixed positive dose, individual replies can remain enthusiastic. For example, a $+0.3$ greedy reply to a new-job announcement begins ``OMG that's insane''. Human ties and the coherence costs likewise show that the intervention changes a tendency rather than deterministically selecting a personality.

\section{Limitations}
\label{sec:limits}
We tested how adjusting L9N7515's activation changes the warmth and interpersonal distance of Qwen3.5-4B's replies, using a fixed character prompt and short conversations. The findings describe this setting, and evaluation across additional models, character roles, and longer dialogues would clarify how broadly the effect transfers. The small prompt sets and dependence between dialogue turns also favor interpreting the reported averages as descriptive results.

Agreement across three model raters provides a consistency check, although the raters assess the same outputs and may share evaluation biases. Human comparisons offer additional evidence for the single-message distance effect.

Steering sometimes reduced task engagement or led the character to step out of role. The firmer character prompt reduced role breaks, while dose selection allowed a balance between distance and coherence. Preserving task completion and character consistency across a wider range of messages remains a useful direction for refinement. Persona labels here describe the style of generated replies.

Candidate and dose selection were exploratory, and the comparisons focus on unsteered and dose-matched random-neuron controls. Comparisons with dense activation steering and explicit personality prompts would help assess relative effectiveness and selection efficiency. Dose values also depend on the reference passage used for residual calibration, so comparisons across implementations require matching that calibration.

\section{Conclusion}
We identified L9N7515 in Qwen3.5-4B by comparing activations on guarded and eager replies written in the same character voice. Adjusting this neuron's activation shifted the tested character's replies between warmer and more distant styles without changing model weights. Positive doses increased rated interpersonal distance beyond the tested random-neuron controls, while message-level changes in intervention direction produced corresponding shifts within short dialogues. Agreement across three model raters and a blind human comparison supports the observed distance effect.

These findings motivate targeted activation steering as a route toward persona customization and task-specific model adaptation without additional fine-tuning. Selecting and adjusting a small set of activations at inference time could reduce the need for repeated training and offer a lightweight way to adapt a shared model to different interaction styles. This creates an opportunity to reduce adaptation-related computation, resource use, and implementation effort. Future work can assess these potential savings against fine-tuning and prompting baselines, extend the method to task-specific behaviors, and improve task engagement and character consistency.

\section*{Acknowledgment}
The authors acknowledge the use of AI-based tools to assist in drafting and editing the text. All content was reviewed and verified by the authors, who take full responsibility for this work.

\section*{Code Availability}
The code for steering and evaluation, together with all blind ratings and the human comparison data, is available from the authors on request.

\end{document}